\documentclass[10pt,twocolumn,letterpaper]{article}
\usepackage[pagenumbers]{cvpr}
\usepackage[T1]{fontenc}
\usepackage{placeins}
\usepackage{stfloats}
\usepackage{amsmath,amssymb,graphicx,booktabs,xcolor}
\definecolor{cvprblue}{rgb}{0.21,0.49,0.74}
\usepackage[pagebackref,breaklinks,colorlinks,allcolors=cvprblue]{hyperref}
\renewcommand*{\backrefalt}[4]{}
\def\confName{}
\def\confYear{}
\title{EAST: Event-Anchored Surface Transport for Event-Based Gaussian Reconstruction}
\author{Ke Luo\textsuperscript{1} \quad Wei Zhai\textsuperscript{1,*} \quad Yuliang Wu\textsuperscript{1} \quad Pengfei Yu\textsuperscript{2} \quad Yang Cao\textsuperscript{1,*}\\
\textsuperscript{1}University of Science and Technology of China \quad \textsuperscript{2}Li Auto\\
{\small\texttt{luoke@mail.ustc.edu.cn}, \texttt{wzhai056@ustc.edu.cn}, \texttt{tronliang@mail.ustc.edu.cn}}\\
{\small\texttt{yupengfei1@lixiang.com}, \texttt{forrest@ustc.edu.cn}}\\
\textsuperscript{*}Corresponding authors}

\hypersetup{pdfauthor={Ke Luo, Wei Zhai, Yuliang Wu, Pengfei Yu, Yang Cao}}
\begin{document}
\maketitle
\flushbottom
\begin{abstract}
Event-supervised Gaussian reconstruction combines the high temporal resolution and wide dynamic range of event cameras with efficient explicit rendering. However, recovering a continuous surface remains difficult: changes in rendered appearance can compensate for depth errors while preserving the predicted brightness changes. We introduce Event-Anchored Surface Transport (EAST), which couples an event-constrained signed distance field to Gaussian rendering. Triangulation-supported candidates anchor the zero set, while normal flow measured from event timing constrains ray--surface intersection depths. Gaussian centers receive bounded corrections from the descriptor difference between the live field and a frozen initial reference. The correction is zero at initialization, yet transmits image gradients to the live field. Experiments on PAEv3D and NeRF Synthetic show improved surface accuracy and reference-view rendering over Gaussian baselines, together with efficient mesh extraction from the learned field.
\end{abstract}

\section{Introduction}
Event cameras record asynchronous brightness changes with high temporal resolution and a wide dynamic range~\cite{gallego2022survey}, supporting reconstruction under rapid motion and challenging illumination. Event-supervised Gaussian methods fit these measurements through differentiable rasterization and render the learned scene efficiently~\cite{kerbl2023gaussians,event3dgs,huang2025inceventgs,yura2025eventsplat}. Recovering a mesh requires the primitives that explain image changes to also determine a consistent three-dimensional surface~\cite{sugar,huang2024twodgs}.

RGB images directly constrain intensity and texture at each observed viewpoint. Events instead record thresholded changes in log intensity, leaving absolute appearance unobserved~\cite{gallego2022survey}. Event-based radiance fields recover appearance and density by explaining this temporal signal~\cite{rudnev2022eventnerf,klenk2023enerf,low2023robustenerf}. Surface reconstruction must additionally determine where along the moving rays those changes originate.

This difference makes appearance--depth compensation particularly relevant to event supervision. Under local translation, the predicted brightness-change response is proportional to image-gradient magnitude and inverse depth. A farther edge with a stronger gradient can produce the same first-order change as a nearer, weaker edge. Gaussian appearance, opacity, and projected footprint can adjust this gradient without correcting depth. Without direct intensity and texture observations to constrain these appearance adjustments, event-only fitting requires geometric evidence beyond brightness-change agreement.

\begin{figure*}[t]
\centering
\includegraphics[width=\linewidth]{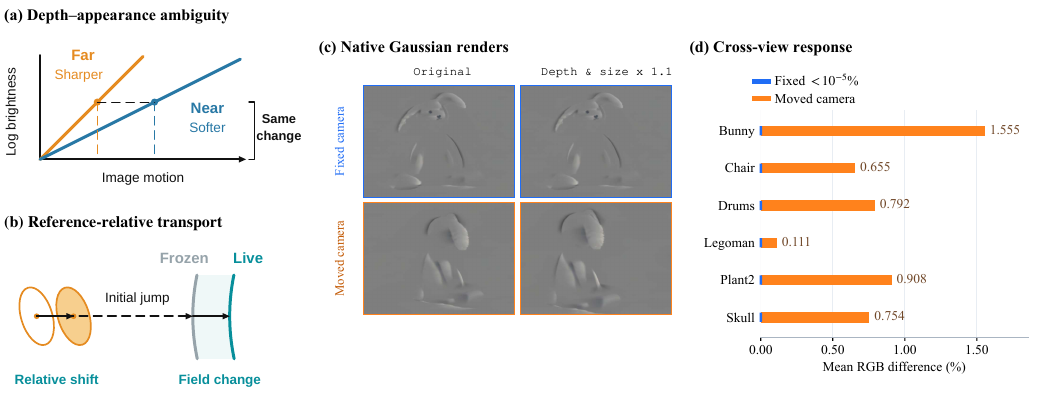}
\caption{\textbf{Depth ambiguity and reference-relative coupling.} (a) Depth and appearance gradient can compensate in local brightness change. (b) Transport follows the live--reference field difference, giving zero correction at initialization. (c) Native Bunny renders before and after radial depth-and-size scaling by $1.1$, with colors and opacity fixed. Rows use fixed and translated cameras. (d) Full-frame RGB MAE multiplied by 100 for six event-trained Gaussian banks. Fixed-camera differences are below $10^{-5}\%$; translation reveals the changed geometry. The supplement details the intervention.}
\label{fig:ambiguity}
\end{figure*}

Figure~\ref{fig:ambiguity}(a,c,d) connects the local ambiguity to a controlled rendering diagnostic. Scaling Gaussian centers about a fixed camera and scaling their covariances accordingly preserves projected footprints under unchanged visibility, while a translated view changes the rendered image. This diagnostic motivates constraints that connect surface location across viewpoints.

Triangulation and normal flow provide complementary geometric evidence. Triangulation tests whether rays from distinct camera poses support a common spatial location, supplying positional constraints across views~\cite{rebecq2018emvs,zhou2018stereoevent}. Normal flow describes how observed edge structure moves in the image over time. Through calibrated camera motion, it constrains depth along viewing rays without using the image gradient chosen by the renderer~\cite{hui2015normal}. Its information is directional and depends on motion relative to the edge, so it complements spatial ray agreement. An implicit field supplies a shared surface~\cite{yariv2020idr,wang2021neus,eventneus}, allowing measurements from different positions, times, and edge orientations to constrain the same boundary.

An SDF alone does not resolve appearance--depth compensation; accurate surface placement requires geometric constraints and effective rendering--surface coupling~\cite{eventneus,yu2024gsdf}. Coupling a surface field to a Gaussian renderer~\cite{3dgsr,surfacesplat} must reconcile their geometric estimates. Moving primitives directly onto the current zero set would transfer this mismatch to rendering. Their coupling must therefore express geometric refinement relative to the initial field (Fig.~\ref{fig:ambiguity}(b)), so that surface changes affect the images without imposing an initial relocation. Returning image residuals through the same correction lets rendering refine the shared boundary.

We propose Event-Anchored Surface Transport (EAST). Calibrated event rays provide three-dimensional candidates, and local event timing provides edge-normal motion. These observations constrain a signed distance field (SDF) through soft zero-set anchors and ray-intersection depths. To couple the field to rendering, EAST evaluates the live field and a frozen initial reference at each current Gaussian center. Their descriptor difference produces a bounded correction that is initially zero but remains differentiable with respect to the live field. Geometric observations constrain the SDF directly, and event-image residuals refine it through the transported centers.

Our contributions are:
\begin{itemize}
\item A local analysis of appearance-gradient--depth compensation that explains how measured event motion adds depth information to brightness-change fitting.
\item Reference-relative surface transport that combines supported SDF anchors and measured normal flow with Gaussian rendering, enabling geometry and image residuals to refine a shared surface.
\item Improved surface accuracy and rendering over event-supervised Gaussian baselines on PAEv3D and NeRF Synthetic, with efficient native surface extraction.
\end{itemize}

\section{Related Work}
\paragraph{Event-based fields and geometry.}
Event multi-view stereo locates scene structure through agreement among calibrated event rays. EMVS accumulates back-projected rays and identifies regions of high ray density~\cite{rebecq2018emvs}, while stereo event reconstruction exploits observations from two event cameras~\cite{zhou2018stereoevent}. Contrast maximization estimates motion, depth, and optical flow by aligning warped events~\cite{gallego2018contrast}. These geometric constructions recover structure around event-generating edges. Implicit fields combine such localized evidence across views. Neural rendering learns a scene representation whose predicted brightness changes explain the event stream. EventNeRF~\cite{rudnev2022eventnerf}, E-NeRF~\cite{klenk2023enerf}, and Robust e-NeRF~\cite{low2023robustenerf} learn radiance and density fields, enabling novel-view synthesis from event supervision. Radiance fields integrate along rays; surface fields locate a shared boundary. IDR~\cite{yariv2020idr} jointly learns an implicit surface and its appearance from multi-view images. NeuS~\cite{wang2021neus} connects an SDF to volume rendering, and EventNeuS~\cite{eventneus} combines SDF and density-field learning under event supervision. A zero set supplies a continuous surface representation; calibrated ray agreement and measured edge motion supply observations of its location.

\paragraph{Event-supervised Gaussian reconstruction.}
Event-3DGS~\cite{event3dgs} and E-3DGS~\cite{edgs} adapt explicit Gaussian rendering to event observations, with E-3DGS addressing large and unbounded scenes. IncEventGS~\cite{huang2025inceventgs} reconstructs incrementally through tracking and mapping: the current Gaussian scene supports camera-motion estimation, and mapping jointly refines the scene and trajectory. EventSplat~\cite{yura2025eventsplat} uses event-to-video initialization and spline-interpolated poses to reconstruct from moving event cameras. EvGGS studies generalizable Gaussian reconstruction~\cite{wang2024evggs}, Elite-EvGS distills event-to-video priors~\cite{zhang2025elite}, and E2EGS addresses pose-free reconstruction~\cite{kim2026e2egs}. GPERT~\cite{gpert} further separates per-event depth ray tracing from snapshot intensity rendering, using warped events to connect geometry and photometry at different temporal resolutions. EAST couples this image-fitting capacity to an event-constrained surface.

\paragraph{Interfaces between Gaussians and surfaces.}
Gaussian surface methods differ in how primitives represent geometry and how they exchange information with an implicit field. SuGaR~\cite{sugar} regularizes Gaussians toward surface alignment for mesh extraction. 2D Gaussian Splatting~\cite{huang2024twodgs} uses oriented planar disks and perspective-correct splatting, giving the primitives an explicit surface interpretation. Geometry Field Splatting~\cite{jiang2025geometry} parameterizes a geometry field with Gaussian surfels and converts it to density for rendering. Gaussian Opacity Fields~\cite{yu2024gof} extracts surfaces as opacity level sets with adaptive tetrahedral meshing. Hybrid methods retain distinct Gaussian and implicit representations. NeuSG~\cite{neusg} uses regularized Gaussian geometry to guide implicit reconstruction, with surface normals helping refine the Gaussian points. 3DGSR~\cite{3dgsr} links SDF values to Gaussian opacity and aligns rendered geometric attributes. GSDF~\cite{yu2024gsdf} uses mutual guidance between Gaussian and SDF branches, while SurfaceSplat~\cite{surfacesplat} initializes Gaussians from an SDF mesh and uses rendered novel views to refine the implicit surface. EAST instead transports Gaussian centers through bounded descriptor changes relative to a frozen field. This preserves the initialized renderer and carries image gradients to the event-constrained SDF.

\section{Event-Anchored Surface Transport}
\begin{figure*}[!t]
\centering
\includegraphics[width=\linewidth]{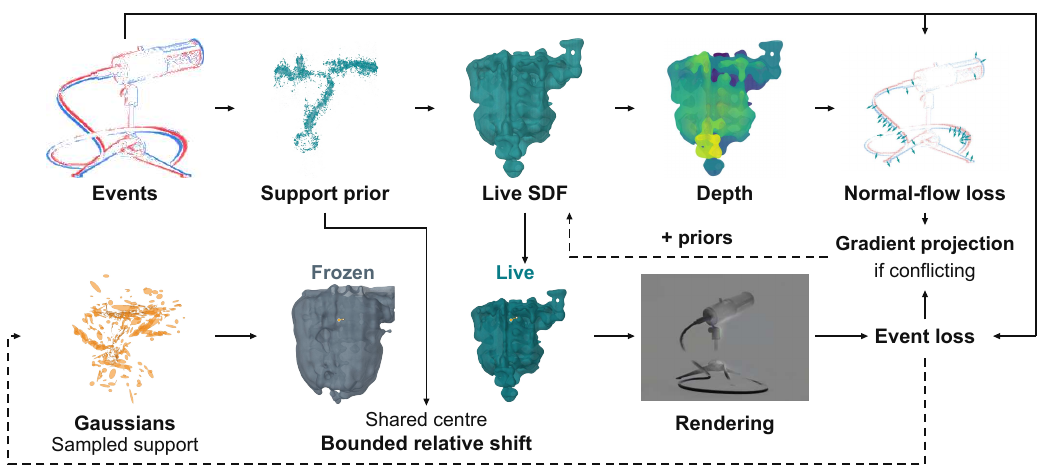}
\caption{\textbf{EAST framework.} Event-supported candidates and measured normal flow constrain a live SDF. Relative to a frozen reference, its descriptor changes produce bounded Gaussian-center displacements. Rendering residuals return through these centers to refine the field; conflicting image and motion gradients are projected before the update.}
\label{fig:overview}
\end{figure*}

EAST couples a continuous surface to Gaussian rendering through event-derived geometry and reference-relative center displacements (Fig.~\ref{fig:overview}).

\subsection{Representation and Geometric Evidence}
\label{sec:representation}
\label{sec:ambiguity}
The inputs are events $e_k=(u_k,t_k,p_k)$, intrinsics $K$, and calibrated poses, where $u_k$, $t_k$, and $p_k\in\{-1,+1\}$ denote pixel, time, and polarity. A signed distance field $f_\theta:\mathbb R^3\to\mathbb R$ defines the surface $\mathcal S_\theta=\{x:f_\theta(x)=0\}$. Gaussian parameters $\phi$ comprise trainable centers $\mu_i$, scales, rotations, opacities, and appearance coefficients. Standard front-to-back rasterization~\cite{kerbl2023gaussians,edgs} uses transported centers $\widetilde\mu_i$ to render intensity $Y(u,t)$.

For window $w=(t_a,t_b)$ and contrast threshold $C$, the predicted signed event count is
\begin{equation}
\widehat n_w(u)=\frac{\log(Y(u,t_b)+\varepsilon_I)-\log(Y(u,t_a)+\varepsilon_I)}{C}.
\label{eq:event_prediction}
\end{equation}
The observed count $n_w(u)$ sums event polarities. Its depth information is coupled to rendered appearance: with $L=\log Y$, calibrated motion gives
\begin{equation}
\partial_tL=-\nabla_uL^\top\dot u,\qquad
\dot u=Z^{-1}A(u)\mathbf v+B(u)\boldsymbol\omega,
\label{eq:flow}
\end{equation}
where $Z$ is camera-axis depth, $\mathbf v,\boldsymbol\omega$ are camera velocities, and $A,B$ are calibrated projection matrices~\cite{hui2015normal}. Under local translation, $q=|\partial_tL|\simeq\gamma g/Z$, with image-gradient magnitude $g$ and fixed factor $\gamma$. Proportional changes $\delta g/g=\delta Z/Z$ preserve $q$ to first order. Measured edge-normal motion constrains depth without depending on the adjustable rendered gradient. Its inverse-depth sensitivity is $\mathbf n^\top A(u)\mathbf v$ for measured unit normal $\mathbf n$; pure rotation and translation tangent to the edge give zero sensitivity. Triangulated positions complement this directional constraint. For inverse depth $\rho=1/Z$ and translational sensitivity $\alpha=\mathbf n^\top A(u)\mathbf v$, measured normal velocity is $\nu=\alpha\rho$. The joint sensitivity of brightness response and measured motion to appearance and inverse depth is
\begin{equation}
\frac{\partial(q,\nu)}{\partial(g,\rho)}=
\begin{pmatrix}\gamma\rho&\gamma g\\0&\alpha\end{pmatrix}.
\label{eq:sensitivity}
\end{equation}
Its determinant is $\gamma\rho\alpha$. At finite positive depth and $\gamma\alpha\ne0$, the two observations locally separate appearance-gradient magnitude from depth. Normal flow therefore supplies a geometric constraint complementary to brightness-change fitting. The supplement gives the derivation and projection diagnostic.

\subsection{Supported Anchoring}
\label{sec:anchoring}
Temporal stereo tests depths along calibrated event rays. Synchronized event-map responses, compared with shifted pose or event-map controls, give candidate positions $x_j$ and confidences $c_j$. A support count $s_j$ records accepted ray-pair triangulations with positive depths, sufficient angular separation, and nearby closest points. The count is set to zero if fewer than two proposals pass or their depth dispersion fails the consistency test. Supported depths combine the original estimate and median accepted proposal. These positions and their evidence remain fixed during joint refinement.

For reliability $r_j=c_j/\max(\max_k c_k,\varepsilon_a)$ and sampled candidates $\mathcal B_a$, the zero-set loss is
\begin{equation}
\mathcal L_{\mathrm{anchor}}=
\frac{\sum_{j\in\mathcal B_a}r_j\mathbf1[s_j>0]f_\theta(x_j)^2}
{\max(\sum_{j\in\mathcal B_a}r_j\mathbf1[s_j>0],\varepsilon_a)}.
\label{eq:anchor}
\end{equation}
Candidate neighborhoods also determine where the field affects rendering. For regularized distance $q_j(x)$ normalized by radius $\rho_j$, define
\begin{equation}
w(x)=1-\prod_j[1-r_j\kappa(q_j(x))],\quad
\kappa(q)=(1-q)_+^4(4q+1).
\label{eq:support}
\end{equation}
With $\mu_i^0$ denoting the center at joint initialization, each bound Gaussian receives fixed weight $w_i=w(\mu_i^0)$ and transport radius $\ell_i$ from its nearest candidate. Anchor selection uses positive triangulation support; transport neighborhoods use the selected confidence-weighted candidates, including those outside the anchor subset. Unbound Gaussians have $w_i=0$. The supplement specifies candidate selection and radii.

\subsection{Measured Normal Flow}
\label{sec:normal_flow}
An affine time plane $T_k(u)=a_k^\top u+\beta_k$ is fitted to causal same-polarity events. A coherent moving edge satisfies $a_k^\top\dot u=1$, yielding measured normal $a_k/\|a_k\|$ and speed $1/\|a_k\|$. Substitution into Eq.~\eqref{eq:flow} gives
\begin{equation}
r_k^{\mathrm{nf}}=A_k/z_k-b_k,\quad
\begin{cases}
A_k=a_k^\top A(u_k)\mathbf v_k,\\
b_k=1-a_k^\top B(u_k)\boldsymbol\omega_k,
\end{cases}
\label{eq:nf}
\end{equation}
where $z_k$ is camera-axis depth at an SDF ray intersection. The measured coefficients remain fixed as the field changes. We refine the first positive-to-nonpositive crossing along a ray $x(s)=o+sd$ inside the reconstruction box. For an accepted transverse root, implicit differentiation gives
\begin{equation}
\frac{\partial s}{\partial\theta}
=-\frac{\partial_\theta f_\theta(o+sd)}{\nabla f_\theta(o+sd)^\top d}.
\label{eq:root_derivative}
\end{equation}
The camera transform from $s$ to $z_k$ propagates the residual to the SDF. With validity $m_k$ and fixed initial invalid-root penalty $\eta_k^0$, the batch loss is
\begin{equation}
\mathcal L_{\mathrm{nf}}=\frac1{|\mathcal B_n|}\sum_{k\in\mathcal B_n}
\begin{cases}(r_k^{\mathrm{nf}})^2,&m_k=1,\\\eta_k^0,&m_k=0.\end{cases}
\label{eq:loss_nf}
\end{equation}
Invalid roots contribute no depth gradient; the batch denominator stays fixed. The supplement gives fitting, validity, and penalty definitions.

\subsection{Reference-Relative Transport}
\label{sec:transport}
Linearizing the zero set gives the stabilized normal correction
\begin{equation}
d_\theta(x)=-\frac{f_\theta(x)\nabla f_\theta(x)}{\max(\|\nabla f_\theta(x)\|^2,\varepsilon_d)}.
\label{eq:descriptor}
\end{equation}
With an inactive denominator floor, this is the minimum-norm displacement to the local zero plane. Direct projection can transfer the mismatch between independently initialized fields and Gaussians. EAST freezes the initial field as $f_0$ and evaluates both descriptors at the same current center:
\begin{equation}
\Delta_i=d_\theta(\mu_i)-d_0(\mu_i),\qquad
\widetilde\mu_i=\mu_i+w_iC_{\ell_i}(\Delta_i),
\label{eq:transport}
\end{equation}
where $C_\ell(a)=a\min(1,\ell/\max(\|a\|,\varepsilon_d))$. Thus $\|\widetilde\mu_i-\mu_i\|\leq w_i\ell_i$: the cap bounds the added correction from the current trainable center, while the base Gaussian parameters remain trainable.

When $f_\theta=f_0$, the descriptors cancel at every query and $\widetilde\mu_i=\mu_i$. This identity preserves the initial renderer while retaining a gradient to the live field. For an inactive cap,
\begin{equation}
\nabla_\theta\mathcal L_{\mathrm{photo}}
=\sum_i w_i\left[\partial_\theta d_\theta(\mu_i)\right]^\top
\nabla_{\widetilde\mu_i}\mathcal L_{\mathrm{photo}}.
\label{eq:gradient}
\end{equation}
With the denominator floor inactive before and after rescaling, multiplying $f$ by a constant $a\ne0$ leaves $d$ unchanged: its numerator and denominator both scale by $a^2$. The descriptor's geometric meaning is therefore independent of field amplitude in this regime. For parallel planes $f_0(x)=\mathbf n^\top x-b_0$ and $f_\theta(x)=\mathbf n^\top x-b_\theta$, with $\|\mathbf n\|=1$, their descriptor difference is $\Delta=(b_\theta-b_0)\mathbf n$. The center correction follows this plane-offset change with the support weight and cap in Eq.~\eqref{eq:transport}. In the unit-weight, uncapped case, $\mathbf n^\top\widetilde\mu_i-b_\theta=\mathbf n^\top\mu_i-b_0$. Thus the transported center retains the current center's signed reference-plane offset as its signed live-plane offset, transferring the plane's motion through a relative change.

Only the live descriptor depends on $\theta$; both fields retain derivatives with respect to their common query $\mu_i$. The supplement derives the center Jacobian and local-plane behavior. The feedback ablation detaches this return gradient.

\subsection{Joint Objective}
\label{sec:optimization}
Event-count fitting over windows $\mathcal W$ and pixels $\Omega$ uses
\begin{equation}
\mathcal L_{\mathrm{photo}}=\frac{\sum_{w,u}|\widehat n_w(u)-n_w(u)|}{|\mathcal W||\Omega|}
+\lambda_g\mathcal L_{\mathrm{gauge}}.
\label{eq:loss_photo}
\end{equation}
The gauge aligns spatial channel means with an internally reconstructed grayscale event target, fixing the log-intensity offset. Eikonal regularization $\mathcal L_{\mathrm{eik}}=\mathbb E_x(\|\nabla f_\theta(x)\|_2-1)^2$ samples uniform and candidate-centered locations. The objective is
\begin{equation}
\mathcal L=\mathcal L_{\mathrm{photo}}+\lambda_a\mathcal L_{\mathrm{anchor}}
+\lambda_n\mathcal L_{\mathrm{nf}}+\lambda_e\mathcal L_{\mathrm{eik}}.
\label{eq:joint_objective}
\end{equation}
For $g=\nabla_\theta\mathcal L_{\mathrm{photo}}$ and $h=\nabla_\theta(\lambda_n\mathcal L_{\mathrm{nf}})$, symmetric gradient projection~\cite{yu2020pcgrad} resolves conflicts. If $g^\top h<0$ and both norms are nonzero,
\begin{equation}
g'=g-\frac{g^\top h}{\|h\|^2}h,\qquad
h'=h-\frac{g^\top h}{\|g\|^2}g.
\label{eq:conflict_projection}
\end{equation}
Both projections use the original pair; otherwise the gradients are unchanged. Let $v=g'+h'$. For a conflicting pair, Cauchy--Schwarz gives
\begin{equation}
g^\top v=\|g\|^2-\frac{(g^\top h)^2}{\|h\|^2}\geq0,
\label{eq:nonascent}
\end{equation}
and analogously $h^\top v\geq0$. Hence $-v$ is a non-ascent direction for both local loss linearizations. This property applies to the projected pair alone. The SDF update adds anchor and Eikonal gradients to $v$ before applying Adam; Gaussian parameters receive the photometric gradient. Initialization, schedules, and evaluation settings appear in Sec.~4; full optimization details are in the supplement.

\section{Experiments}
\suppressfloats[t]
\newcommand{\PAEComparisonTable}{%
\begin{table*}[!b]
\centering\small
\caption{\textbf{PAEv3D reconstruction under the 30K schedule.} Means over six objects. Geometry uses world units; RGB scores use reference views. Best and second-best results are bold and underlined.}
\label{tab:pae_compare}
\setlength{\tabcolsep}{3.5pt}
\renewcommand{\arraystretch}{1.16}
\begin{tabular*}{\linewidth}{@{\extracolsep{\fill}}llrrrrrr@{}}
\toprule
Method & Surface & \multicolumn{3}{c}{Surface geometry $\downarrow$} & \multicolumn{3}{c}{Reference-view rendering} \\
\cmidrule(lr){3-5}\cmidrule(lr){6-8}
 & & CDsum & Acc. & Comp. & PSNR $\uparrow$ & SSIM $\uparrow$ & LPIPS $\downarrow$ \\
\midrule
EventNeRF (CVPR 2023)~\cite{rudnev2022eventnerf} & Density & \underline{0.11} & \underline{0.105} & \textbf{0.005} & 10.391 & 0.607 & 0.518 \\
Event-3DGS (NeurIPS 2024)~\cite{event3dgs} & Depth + TSDF & 0.127 & 0.107 & 0.021 & \underline{13.007} & \underline{0.694} & \underline{0.321} \\
IncEventGS (CVPR 2025)~\cite{huang2025inceventgs} & Depth + TSDF & 0.135 & 0.122 & 0.013 & 10.43 & 0.635 & 0.346 \\
EventNeuS (3DV 2026)~\cite{eventneus} & Native SDF & 0.202 & 0.161 & 0.041 & 12.792 & 0.647 & 0.457 \\
\midrule
\textbf{EAST} & Native SDF & \textbf{0.051} & \textbf{0.045} & \underline{0.006} & \textbf{13.573} & \textbf{0.707} & \textbf{0.306} \\
\bottomrule
\end{tabular*}
\end{table*}
}

\newcommand{\SyntheticComparisonTable}{%
\begin{table*}[!b]
\centering\small
\caption{\textbf{NeRF Synthetic reconstruction under the 30K schedule.} Means over all five scenes. Geometry uses world units; RGB scores use reference views. Best and second-best results are bold and underlined.}
\label{tab:original5}
\setlength{\tabcolsep}{3.5pt}
\renewcommand{\arraystretch}{1.16}
\begin{tabular*}{\linewidth}{@{\extracolsep{\fill}}llrrrrrr@{}}
\toprule
Method & Surface & \multicolumn{3}{c}{Surface geometry $\downarrow$} & \multicolumn{3}{c}{Reference-view rendering} \\
\cmidrule(lr){3-5}\cmidrule(lr){6-8}
 & & CDsum & Acc. & Comp. & PSNR $\uparrow$ & SSIM $\uparrow$ & LPIPS $\downarrow$ \\
\midrule
EventNeRF (CVPR 2023)~\cite{rudnev2022eventnerf} & Density & \textbf{0.026} & \textbf{0.018} & \underline{0.008} & 16.655 & 0.782 & 0.431 \\
Event-3DGS (NeurIPS 2024)~\cite{event3dgs} & Depth + TSDF & 0.083 & 0.067 & 0.016 & \underline{17.383} & \underline{0.843} & \underline{0.266} \\
IncEventGS (CVPR 2025)~\cite{huang2025inceventgs} & Depth + TSDF & 0.051 & 0.041 & 0.01 & 14.098 & 0.786 & 0.422 \\
EventNeuS (3DV 2026)~\cite{eventneus} & Native SDF & 0.033 & 0.025 & \textbf{0.008} & 16.101 & 0.709 & 0.528 \\
\midrule
\textbf{EAST} & Native SDF & \underline{0.028} & \underline{0.019} & 0.009 & \textbf{17.551} & \textbf{0.881} & \textbf{0.234} \\
\bottomrule
\end{tabular*}
\end{table*}
}

\PAEComparisonTable
\SyntheticComparisonTable

\subsection{Experimental Setup}
We evaluate EAST on six PAEv3D objects and five NeRF Synthetic scenes: Chair, Drums, Hotdog, Lego, and Mic~\cite{eventneus,mildenhall2020nerf}. Baselines include EventNeRF~\cite{rudnev2022eventnerf}, Event-3DGS~\cite{event3dgs}, IncEventGS~\cite{huang2025inceventgs}, and EventNeuS~\cite{eventneus}. All methods use events, calibration, and supplied poses, without reference RGB supervision or pretrained depth priors.

\paragraph{Initialization and training.}
Gaussians and the SDF each undergo 10K initialization updates, followed by 20K joint updates. Gaussian initialization fits event-derived intensities and event counts; SDF initialization establishes the reference field. Joint training optimizes the live SDF and Gaussians with the reference field, Gaussian count, support weights, and transport radii fixed. Checkpoints are selected by training event loss. Ablations share the same initialization and training schedule. Architecture, hyperparameters, and sampling details appear in the supplement.

\paragraph{Evaluation.}
Accuracy (Acc.) and completeness (Comp.) measure reconstructed-to-reference and reference-to-reconstructed surface distances; CDsum is their sum. Surfaces are extracted on a common $300^3$ grid and compared using 100K area samples in the supplied world coordinates. EAST and EventNeuS extract SDF zero sets, EventNeRF extracts a density isosurface, and Gaussian baselines use depth fusion. Full-frame PSNR, SSIM~\cite{wang2004ssim}, and LPIPS~\cite{zhang2018lpips} are evaluated over 21 reference views per PAEv3D object and seven per Synthetic scene.

\begin{figure*}[t]
\centering
\includegraphics[width=\linewidth]{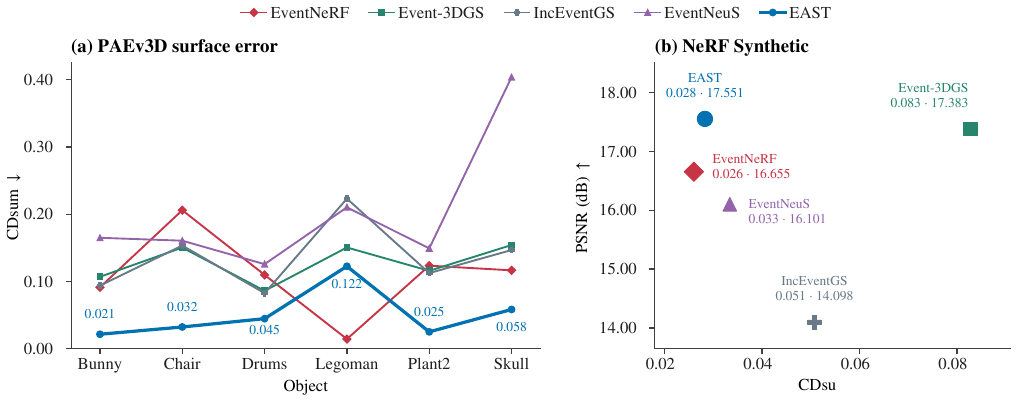}
\caption{\textbf{Geometry and rendering under the 30K schedule.} (a) Object-wise PAEv3D CDsum. (b) NeRF Synthetic mean CDsum and PSNR. Methods: EventNeRF (CVPR 2023)~\cite{rudnev2022eventnerf}, Event-3DGS (NeurIPS 2024)~\cite{event3dgs}, IncEventGS (CVPR 2025)~\cite{huang2025inceventgs}, EventNeuS (3DV 2026)~\cite{eventneus}, and EAST. Geometry uses world units.}
\label{fig:pae_scene_gain}
\end{figure*}

\subsection{Quantitative Reconstruction}
\paragraph{PAEv3D.}
EAST achieves the lowest mean CDsum and accuracy error, together with the best PSNR, SSIM, and LPIPS (Table~\ref{tab:pae_compare}). It leads CDsum on five of six objects (Fig.~\ref{fig:pae_scene_gain}(a)); EventNeRF leads on Legoman and mean completeness. The accuracy gain shows that coupling event-derived geometry to rendering places the reconstructed surface closer to the reference while improving image quality over Gaussian baselines.

\paragraph{NeRF Synthetic.}
EAST achieves the best PSNR, SSIM, and LPIPS, and the second-lowest CDsum after EventNeRF (Table~\ref{tab:original5}, Fig.~\ref{fig:pae_scene_gain}(b)). It improves both surface geometry and image quality over Event-3DGS and IncEventGS. These gains show that the shared SDF strengthens surface reconstruction while retaining the image-fitting capacity of Gaussian rendering.

\subsection{Qualitative Reconstruction}
\begin{figure*}[t]
\centering
\begingroup\setlength{\unitlength}{\linewidth}
\begin{picture}(1,0.412984)
\put(0,0){\includegraphics[width=\linewidth]{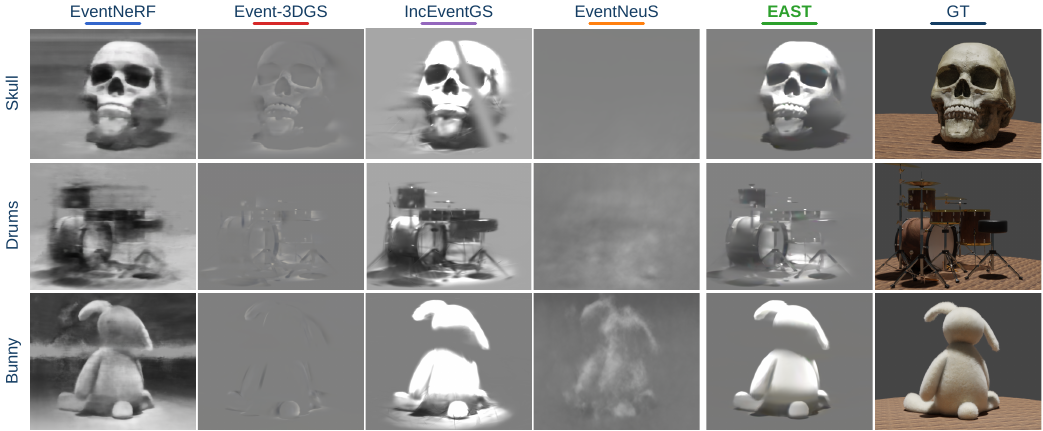}}
\put(0.065116,0.391249){\hyperlink{cite.rudnev2022eventnerf}{\phantom{\rule{0.086619\linewidth}{0.021839\linewidth}}}}
\put(0.222590,0.391249){\hyperlink{cite.event3dgs}{\phantom{\rule{0.093733\linewidth}{0.021839\linewidth}}}}
\put(0.385836,0.391249){\hyperlink{cite.huang2025inceventgs}{\phantom{\rule{0.089305\linewidth}{0.021839\linewidth}}}}
\put(0.549088,0.391249){\hyperlink{cite.eventneus}{\phantom{\rule{0.084860\linewidth}{0.021839\linewidth}}}}
\end{picture}\endgroup
\caption{\textbf{PAEv3D RGB reconstruction.} Matched views of Skull, Drums, and Bunny. Methods: EventNeRF (CVPR 2023)~\cite{rudnev2022eventnerf}, Event-3DGS (NeurIPS 2024)~\cite{event3dgs}, IncEventGS (CVPR 2025)~\cite{huang2025inceventgs}, EventNeuS (3DV 2026)~\cite{eventneus}, and EAST. EAST preserves facial cavities, thin drum supports, and separated feet.}
\label{fig:pae_qualitative}
\end{figure*}

\paragraph{RGB structure.}
EAST preserves Skull's eye sockets, nasal opening, and jaw contours, which are less distinct in Event-3DGS~\cite{event3dgs} (Fig.~\ref{fig:pae_qualitative}). Its drum shells and thin stands remain separated, where EventNeRF~\cite{rudnev2022eventnerf} blurs the supports and IncEventGS~\cite{huang2025inceventgs} introduces bright streaks. Bunny retains distinct feet and clearer ear and body outlines than EventNeuS~\cite{eventneus}.

\begin{figure*}[t]
\centering
\captionsetup{hypcap=false}
\begingroup\setlength{\unitlength}{\linewidth}
\begin{picture}(1,0.560284)
\put(0,0){\includegraphics[width=\linewidth]{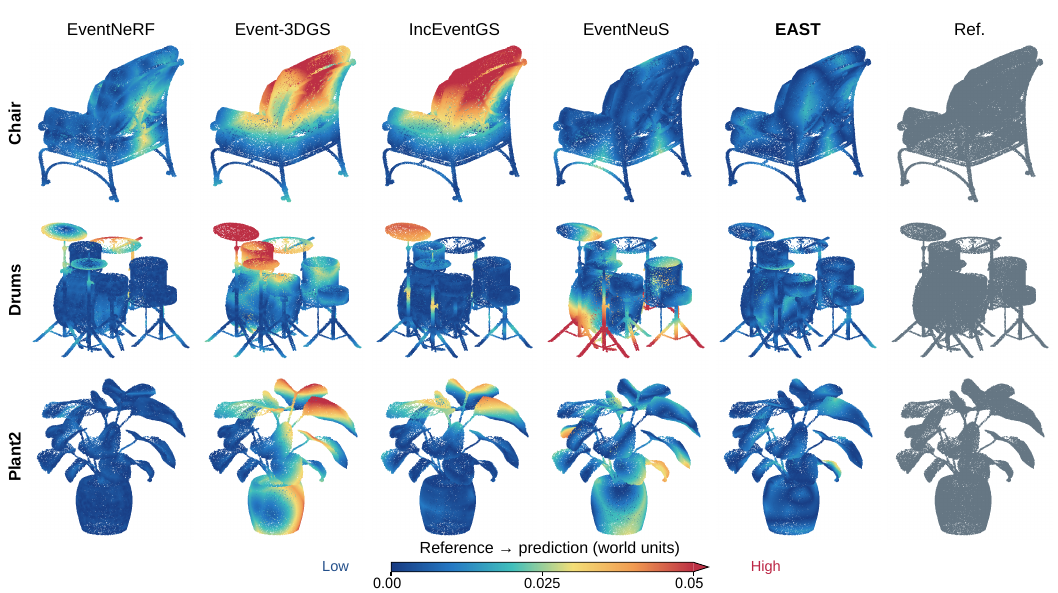}}
\put(0.061283,0.521595){\hyperlink{cite.rudnev2022eventnerf}{\phantom{\rule{0.087426\linewidth}{0.021968\linewidth}}}}
\put(0.220090,0.521595){\hyperlink{cite.event3dgs}{\phantom{\rule{0.094615\linewidth}{0.021968\linewidth}}}}
\put(0.384730,0.521595){\hyperlink{cite.huang2025inceventgs}{\phantom{\rule{0.090140\linewidth}{0.021968\linewidth}}}}
\put(0.549371,0.521595){\hyperlink{cite.eventneus}{\phantom{\rule{0.085649\linewidth}{0.021968\linewidth}}}}
\end{picture}\endgroup
\caption{\textbf{PAEv3D reference-surface error.} EventNeRF (CVPR 2023)~\cite{rudnev2022eventnerf}, Event-3DGS (NeurIPS 2024)~\cite{event3dgs}, IncEventGS (CVPR 2025)~\cite{huang2025inceventgs}, EventNeuS (3DV 2026)~\cite{eventneus}, and EAST. Reference-to-prediction distances use matched geometry and views with a shared 0.00--0.05 world-unit scale (saturated at 0.05); gray denotes the reference.}
\label{fig:cd_complete_reference}
\end{figure*}

\paragraph{Surface error.}
Figure~\ref{fig:cd_complete_reference} maps reference-to-prediction distances. EAST's Chair back and seat are predominantly low-error blue, whereas Event-3DGS~\cite{event3dgs} and IncEventGS~\cite{huang2025inceventgs} show broad warm regions on the upper back. EAST also reduces errors on drum shells and cymbals relative to Event-3DGS and EventNeuS~\cite{eventneus}. On Plant2, most leaves and the pot are blue, with localized deviations on leaves and the pot base.

\begingroup
\interlinepenalty=10000
\subsection{Coupling Ablations}
\begin{table*}[t]
\centering\small
\caption{\textbf{Coupling ablation on NeRF Synthetic at 30K.} Means over all five scenes. Best and second-best results are bold and underlined.}
\label{tab:synthetic_ablation}
\setlength{\tabcolsep}{4pt}
\renewcommand{\arraystretch}{1.40}
\begin{tabular*}{\linewidth}{@{\extracolsep{\fill}}lp{0.40\linewidth}rrr@{}}
\toprule
Configuration & Intervention & CDsum $\downarrow$ & Acc. $\downarrow$ & Comp. $\downarrow$ \\
\midrule
\textbf{EAST} & All components & \textbf{0.028} & \textbf{0.019} & \underline{0.009} \\
w/o support mask & Anchor all candidates & 0.039 & 0.03 & 0.009 \\
w/o rendering feedback & Detach the SDF from image gradients & 0.031 & 0.022 & 0.009 \\
Absolute projection & Use the live descriptor without reference subtraction & \underline{0.029} & \underline{0.02} & \textbf{0.009} \\
w/o gradient projection & Sum photometric and normal-flow gradients & 0.034 & 0.024 & 0.01 \\
w/o NF bundle & Remove the motion loss and its projection interaction & 0.043 & 0.033 & 0.01 \\
\bottomrule
\end{tabular*}
\end{table*}

Table~\ref{tab:synthetic_ablation} reports five-scene means at 30K. The full model achieves the lowest CDsum and accuracy error, supporting the joint use of geometric evidence, relative transport, and rendering feedback.
\par\endgroup

\paragraph{Supported anchors and rendering feedback.}
Anchoring unsupported candidates mainly increases accuracy error. Event-map confidence identifies locations that explain image changes, while triangulation support determines whether calibrated rays establish a surface position. Anchoring candidates without ray agreement introduces uncertain depths into surface supervision. The support mask separates these roles: confidence weights the transport neighborhoods, and ray agreement selects surface anchors.

Removing rendering feedback worsens both directional errors. Supported anchors constrain discrete field locations.

Image gradients further refine the field according to the rendered effect of moving bound Gaussians. Their combined influence couples spatially localized depth evidence with image residuals across views, allowing the shared field to improve both surface placement and coverage.

\paragraph{Relative transport.}
Absolute projection slightly improves completeness but worsens accuracy and CDsum: greater proximity to the reference surface coexists with misplaced predicted geometry. Reference-relative transport better balances surface placement and coverage. It transfers changes of the field's local zero plane to Gaussian centers while preserving their reference-plane offsets in the uncapped planar case (Sec.~\ref{sec:transport}).

\paragraph{Motion supervision and gradient coordination.}
Removing gradient projection increases surface error despite retaining normal-flow supervision, supporting coordination between image and motion updates. Removing the NF bundle, including the motion loss and its projection interaction, causes the largest degradation. Measured motion and coordinated gradients therefore complement supported anchoring and rendering feedback in refining the shared zero set.

\subsection{Efficiency}
EAST extracts Chair in 1.94~s on an RTX 4090, the fastest method in Table~\ref{tab:extraction_efficiency}. Its continuous field provides a surface directly, while Gaussian baselines require multi-view depth fusion.
\par\addvspace{\intextsep}
\noindent\begin{minipage}{\linewidth}
\centering\footnotesize
\captionsetup{hypcap=false}
\captionof{table}{\textbf{Surface extraction efficiency.} Chair on an RTX 4090. Time includes model restoration through PLY export.}
\label{tab:extraction_efficiency}
\setlength{\tabcolsep}{3pt}
\renewcommand{\arraystretch}{1.20}
\begin{tabular*}{\linewidth}{@{\extracolsep{\fill}}llr@{}}
\toprule
Method & Surface & Time (s) $\downarrow$ \\
\midrule
EventNeRF (CVPR 2023)~\cite{rudnev2022eventnerf} & Density & 5.99 \\
Event-3DGS (NeurIPS 2024)~\cite{event3dgs} & Depth TSDF & 15.79 \\
IncEventGS (CVPR 2025)~\cite{huang2025inceventgs} & Depth TSDF & 15.02 \\
EventNeuS (3DV 2026)~\cite{eventneus} & Native SDF & \underline{4.88} \\
\midrule
\textbf{EAST} & Native SDF & \textbf{1.94} \\
\bottomrule
\end{tabular*}
\end{minipage}
\par\addvspace{\intextsep}

\begingroup
\interlinepenalty=10000
\noindent Reusing transported centers across views reduces rendering time from 5.65 to 3.94~ms/frame.
\par\endgroup

\section{Conclusion}
EAST combines supported event-ray anchors and measured normal flow with Gaussian rendering through bounded reference-relative transport. The shared SDF improves surface accuracy and image quality over Gaussian baselines on PAEv3D and NeRF Synthetic, with efficient mesh extraction.

\paragraph{Limitations and future work.}
EAST relies on supplied calibration and camera poses, which determine both triangulation and normal-flow constraints. Normal flow provides depth information only when projected translation has a nonzero component along the measured edge normal; pure rotation and tangential motion must rely on other geometric evidence. The coupling also fixes each Gaussian's support weight and transport radius at initialization, so these bindings cannot adapt as surface estimates change. Joint pose--surface refinement is a promising direction for uncertain trajectories. Adaptive support could follow evolving geometry and incorporate previously unbound Gaussians as new depth evidence becomes available.

\FloatBarrier
\section*{LLM Usage Statement}
The LLM assisted with partial revisions of the paper. The core work, including language presentation, experimental results and methodology design, was independently completed by the authors.

{\small\setlength{\bibsep}{0pt}\bibliographystyle{ieeenat_fullname}\bibliography{references}}
\end{document}